\documentclass[conference]{IEEEtran}
\IEEEoverridecommandlockouts
\renewcommand{\thesection}{\Roman{section}}
\usepackage{booktabs}
\usepackage{fancyhdr}
\usepackage{algorithm}
\usepackage[usenames, dvipsnames]{xcolor}
\usepackage{algorithmic}
\usepackage{colortbl}
\usepackage{caption}
\usepackage{graphicx}
\usepackage{array}
\usepackage{adjustbox}
\usepackage{hyperref,graphicx,color,float}
\definecolor{myblue}{RGB}{10, 150, 200}

\usepackage{cite}
\usepackage{multirow}
\usepackage{amsmath,amssymb,amsfonts}
\usepackage{algorithmic}
\usepackage{graphicx}
\usepackage{textcomp}
\usepackage{comment}
\definecolor{highlightColor}{HTML}{E6FFE6}

\usepackage{xcolor}
\def\BibTeX{{\rm B\kern-.05em{\sc i\kern-.025em b}\kern-.08em
    T\kern-.1667em\lower.7ex\hbox{E}\kern-.125emX}}
    
\title{Edge-Native Digitization of Handwritten Marksheets: A Hybrid Heuristic-Deep Learning Framework}

\author{
    \IEEEauthorblockN{
        Md. Irtiza Hossain \textsuperscript{2}
        Junaid Ahmed Sifat\textsuperscript{1}, 
        Abir Chowdhury\textsuperscript{1}, 
    }
    \IEEEauthorblockA{
        \textsuperscript{1,2}Department of Computer Science and Engineering, Brac University \\
        Dhaka, Bangladesh \\
        Email: 
        \textsuperscript{2}mohammad.irtiza.hossain@gmail.com,
        \textsuperscript{1}junaid.ahmed.sifat@g.bracu.ac.bd,
         \textsuperscript{1}abir.chowdhury1@g.bracu.ac.bd, 
    }
}

\begin{document}
\maketitle

\begin{abstract}
The digitization of structured handwritten documents, such as academic marksheets, remains a significant challenge due to the dual complexity of irregular table structures and diverse handwriting styles. While recent Transformer-based approaches like TableNet and TrOCR achieve state-of-the-art accuracy, their high computational cost renders them unsuitable for resource-constrained edge deployments. This paper introduces a resource-efficient hybrid framework that integrates a heuristic OpenCV-based pipeline for rapid table structure detection with a modified lightweight YOLOv8 architecture for handwritten character recognition. By strategically removing the SPPF and deep C2f layers from the standard YOLOv8 backbone, we reduce computational overhead while maintaining high recognition fidelity. Experimental results on the EMNIST digit benchmark demonstrate that our Modified YOLOv8 model achieves 97.5\% accuracy. Furthermore, we provide a comprehensive efficiency analysis showing that our framework offers a $\approx95\times$ inference speedup over standard OCR pipelines and massive efficiency gains over emerging Large Multimodal Models (LMMs) like Qwen2.5-VL, achieving real-time performance ($\approx$29 FPS) on standard CPU hardware. A qualitative and quantitative evaluation on the \textbf{AMES} dataset, a challenging subset of real-world marksheets, confirms the system's robustness in handling mixed alphanumeric content, bridging the gap between high-performance deep learning and practical, scalable document automation.
\end{abstract}


\section{Introduction}
\label{sec:intro}


The development of the digital database in place of the physical one is fundamental in the current efficiency of administration. Even now when almost everything is digitized, higher education institutions continue to set up tonnes of handwritten marksheets manually. Although there are cloud-based solutions, the use of constant connectivity and high-power infrastructure is still a major bottleneck to the deployment in resource-focused settings \cite{gamage2024optimizing}. This requires two computer vision (CV) challenges including \textit{Table Structure Recognition (TSR)} to segment layouts and \textit{Handwritten Text Recognition (HTR)} to read in order to automate this needs.



Existing solutions are largely based on deep learning heavy models. For TSR, architectures like TableNet \cite{paliwal2019tablenet} and Table Transformers (TATR) \cite{smock2022pubtables} treat this table detection as complex object detection alongside achieving high accuracy but demanding substantial GPU resources \cite{kasem2024deep}. Equally, the most modern HTR models like TrOCR \cite{li2021trocr} utilize large-scale Transformers, which require a high latency, which is not applicable to the work of the typical office computer.

However, on the flip side, when using traditional heuristic methods, there is always a criticism that such algorithms fail to generalize to in-the-wild images, but as of late, lightweight optimizations have been shown to be more effective than giant generalists, when considering a particular edge use case \cite{aminu2025lightweight}. We posit that the use of deep learning in detecting simple grid in academic marksheets is computationally inefficient.

This study provides the link between high-performance deep learning and resource-constrained deployment. We introduce a hybrid framework which is modular, leaving the structural analysis in the hands of a robust yet low-cost heuristic pipeline but with deep learning being the reserve of recognition. Our system combines an optimized table detector based on OpenCV with a new structure of a modified YOLOv8 system. With a carefully crafted set of pruning operations applied to the standard YOLOv8 backbone to remove extra layers the SPPF and deep C2f layers in particular and classify individual characters, we design a light model.

We test our method on the EMNIST benchmark and show that it is effective on a real-life dataset of complex marksheets. The efficiency analysis ensures that this hybrid design offers a real time inference on CPU's and also it provides a scalable solution in case for offline and high volume administrative automation.

\section{Literature Review}
\label{sec:lit_review}

Automated document processing has evolved from rule based heuristics to data driven deep learning. This section briefly reviews the progression in Table Structure Recognition (TSR) and Handwritten Text Recognition (HTR).

\subsection{Table Structure Recognition (TSR)}
The early TSR used heuristic tools such as projection profiles and connected component analysis. Although computationally insignificant, these algorithms have challenges with noise and skew of scanned documents. Contemporary methods consider TSR as a job of identifying objects. Deep learning models like TableNet \cite{paliwal2019tablenet} and the Transformer based TATR \cite{smock2022pubtables} achieve highest accuracy by learning complex spatial dependencies. Nevertheless, they do not fit edge environments due to their high backbone (e.g., ResNet-101, DETR). Other more recent models such as DocLayout-YOLO \cite{zhao2024doclayout} have attempted to adapt YOLO architectures for layout analysis to bridge this gap but still tend to have massive layers of feature aggregation that are not needed when analyzing simple grid structures.

\subsection{Handwritten Text Recognition (HTR)}
HTR has switched to neural sequence modeling as opposed to the use of Hidden Markov Models \cite{beigi1993overview}. Hybrids between CNN and RNN were used as the standard of transcribing sequential text \cite{jebadurai2021handwritten}. Vision Transformers, such as TrOCR \cite{li2021trocr} and Large Multimodal Models (LMMs) such as Qwen2.5-VL \cite{qwen2025vl} have recently established new standards in performance. Though providing excellent arguments to ambiguous handwritings the models have very high computational expenses. Serious contenders such as PP-OCR \cite{du2020pp} try to trade off speed and accuracy but usually need to use the GPU to do inference optimally. 

\subsection{Efficient Object Detection on the Edge}

YOLO family has transformed the real time object detection. YOLOv8 \cite{mupparaju2024review} introduced an anchor free design with a CSPDarknet backbone, significantly improving feature extraction. Yet, the regular YOLO models are to be used with complex multi-object scenes (e.g., COCO). In 2024 and 2025 recent study emphasizes the necessity of Edge Native AI where models are apparently pruned and quantized for low powered devices \cite{aminu2025lightweight}. Deng et al. \cite{deng2023yolo} illustrated that YOLO backbones remain superior for document layout analysis when optimized correctly. We go further by pruning \textit{pruning} our architecture to get rid of layers that we do not need to recognize single character.

\section{Methodology}
\label{sec:method}

\begin{figure*}[htbp]
\centering
\includegraphics[width=\textwidth]{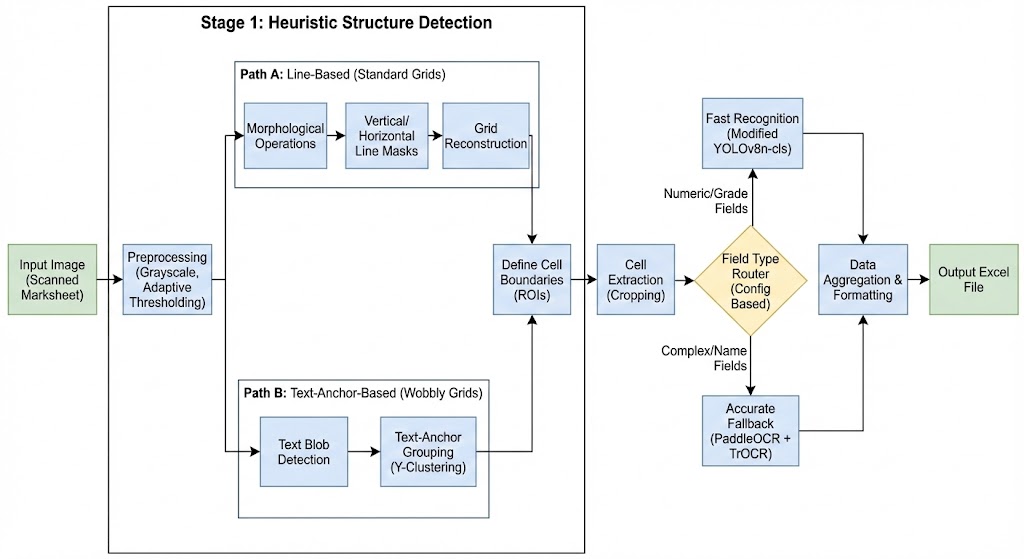}
\caption{The complete workflow of the proposed hybrid system. The pipeline integrates preprocessing, heuristic line detection, cell extraction, and a recognition stage utilizing either Modified YOLOv8 or PaddleOCR.}
\label{fig:workflow}
\end{figure*}

The framework that we propose is a modular hybrid system that will ensure that the throughput of edge devices is maximized and also ensure that the accuracy of the framework is also high when it comes to complex document layouts. The overall workflow is illustrated in Fig. \ref{fig:workflow}.

\subsection{Dataset Preparation}
To ensure robustness against real world variability, we utilized a composite dataset strategy consisting three distinct subsets.

\subsubsection{Our Created Dataset of Handwritten Digits}
\begin{figure}[htbp]
\centering
\includegraphics[width=0.45\textwidth,height=0.20\textheight,keepaspectratio]{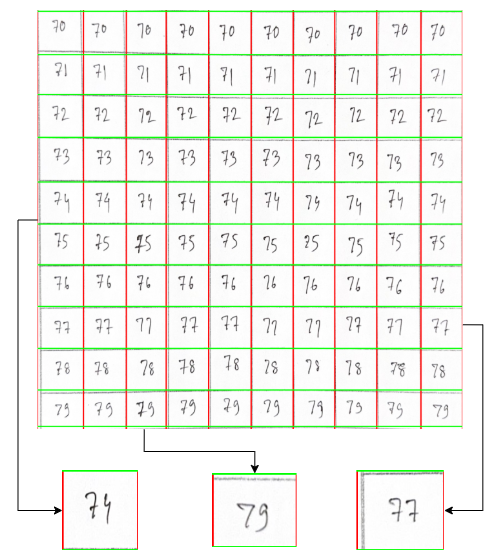}
\caption{Visualization of Our Created Dataset of Handwritten Digits.}
\label{fig:dataset}
\end{figure}

We curated a large scale training dataset consisting of approximately 10,000 handwritten digit and character samples. This dataset combines generated samples such that filling grid templates and collected samples extracted from university documents. This dataset serves as the primary resource for training our recognition models (Fig. \ref{fig:dataset}).

\subsubsection{AMES (Academic Marksheet Evaluation Set)}
To test the truth of this, we presented the AMES dataset, comprising of about 600 full-page images of real marksheets written by hand. Out of this repository, we selected a representative sample to be used as the main reference point. AMES represents the wild distribution of data which features mixed strings of alphanumeric, red colored corrections and irregular grids.

\subsection{Heuristic Table Structure Detection}
\begin{figure*}[htbp]
\centering
\includegraphics[width=\textwidth,height=0.36\textheight,keepaspectratio]{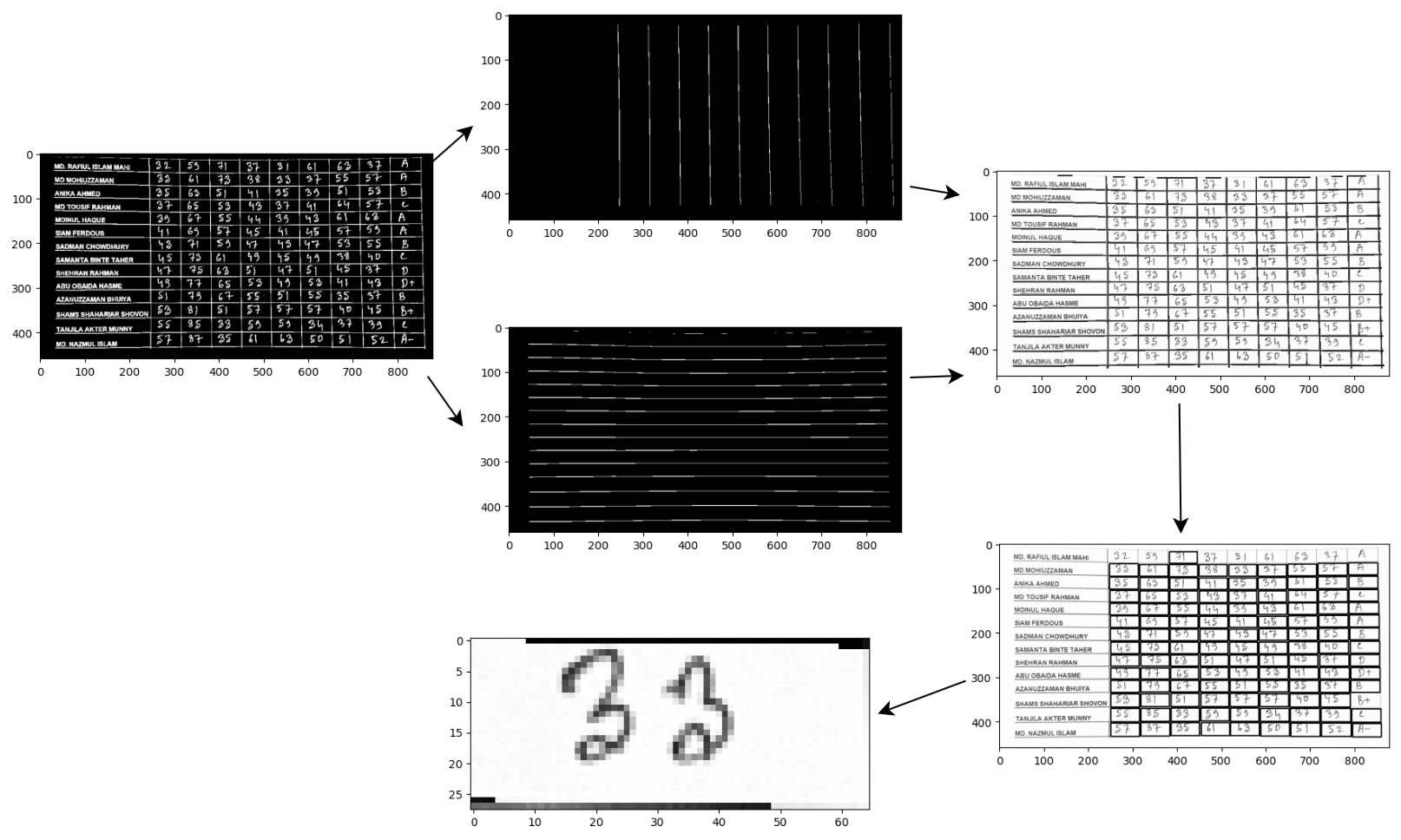}
\caption{Heuristic Table Structure Detection Pipeline.}
\label{fig:structure_detection}
\end{figure*}

We employ a resource based computer vision pipeline by using OpenCv Fig. \ref{fig:structure_detection} to avoid the computational overhead of deep learning based table detectors. It relies on morphological filtering to isolate vertical and horizontal lines and  reconstructing the grid in $O(N)$ pixel time. This plan of action takes advantage of this predictable topology of academic forms so as to prevent the latency of Transformer based inference.

\subsection{Modified YOLOv8 Architecture}
We introduce a Modified YOLOv8 tailored for single character classification. While lightweight classifiers like MobileNet exist, we selected the CSPDarknet backbone to maintain a unified architectural ecosystem.

\begin{figure}[htbp]
\centering
\includegraphics[width=.51\textwidth]{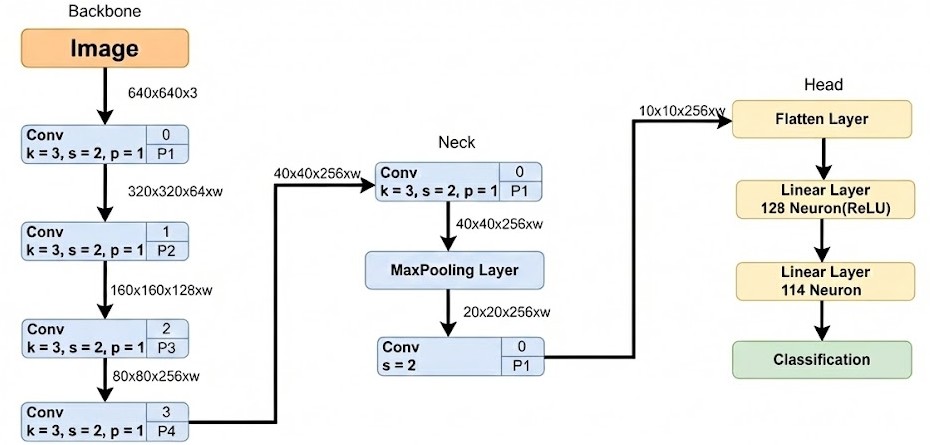}
\caption{Architecture of the Modified YOLOv8. We pruned the SPPF and deep C2f layers from the standard backbone and designed a lightweight classification head with a flattened linear layer, optimizing it for single-character recognition.}
\label{fig:yolo_arch}
\end{figure}

Fig. \ref{fig:yolo_arch} details our architectural modifications:
\begin{itemize}
    \item \textbf{Backbone Pruning:} The Spatial Pyramid Pooling Fast (SPPF) layer was removed, as multi-scale feature capture is unnecessary for pre-cropped characters.
    
    \item \textbf{Neck Optimization:} The depth of the C2f modules was reduced to minimize GFLOPs and parameter count, while retaining the efficient gradient flow of the CSPDarknet.
    
    \item \textbf{Classification Head:} The detection head was replaced with a flattened linear layer, specifically tuned for 10-class digit classification.
\end{itemize}

\subsection{Hybrid Inference Pipeline: Deterministic Routing}
\begin{figure}[htbp]
\centering
\includegraphics[width=0.48\textwidth]{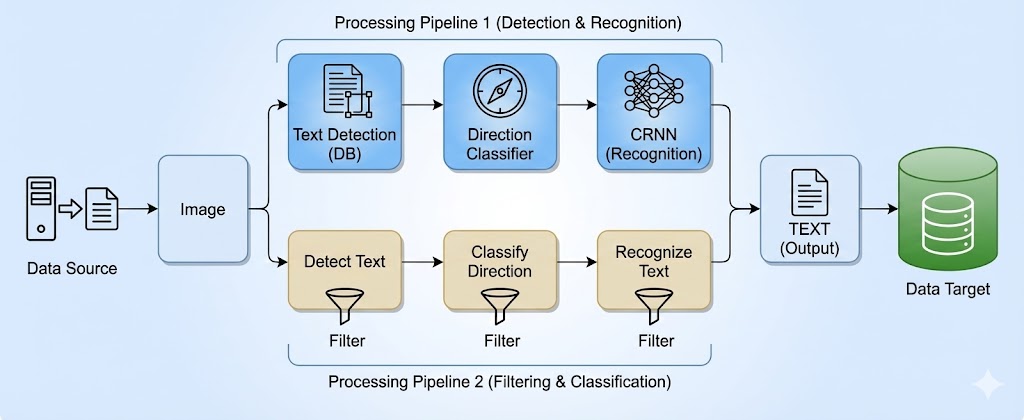}
\caption{PaddleOCR Architecture used for fallback recognition.}
\label{fig:paddle_arch}
\end{figure}

To balance accuracy and speed, we implement a Deterministic Column Routing strategy. Since academic marksheet layouts are standardized, we utilize a template based configuration to route specific columns (e.g., 'Student Name') to the Transformer-based PaddleOCR/TrOCR model (Fig. \ref{fig:paddle_arch}), while numeric columns are processed by the lightweight Modified YOLOv8.


\section{Experiments and Results}
\label{sec:results}

We have compared our hybrid framework along three important dimensions, namely: strong classification behavior in terms of F1-Score measure, fine-grained recognition faithfulness on actual world marksheets, and inference time on resource-limited edge hardware

\subsection{Performance Comparison on EMNIST}
As mentioned earlier, the performance comparison was conducted on the EMNIST.
In order to set the background of isolated character recognition, we tested the modified YOLOv8 classification head on the EMNIST digit dataset. To obtain the harmonic average of precision and recall, instead of using Top-1 Accuracy, which may be distorted, we used the F1 -Score.

\begin{figure}[htbp]
\centering
\includegraphics[width=0.48\textwidth]{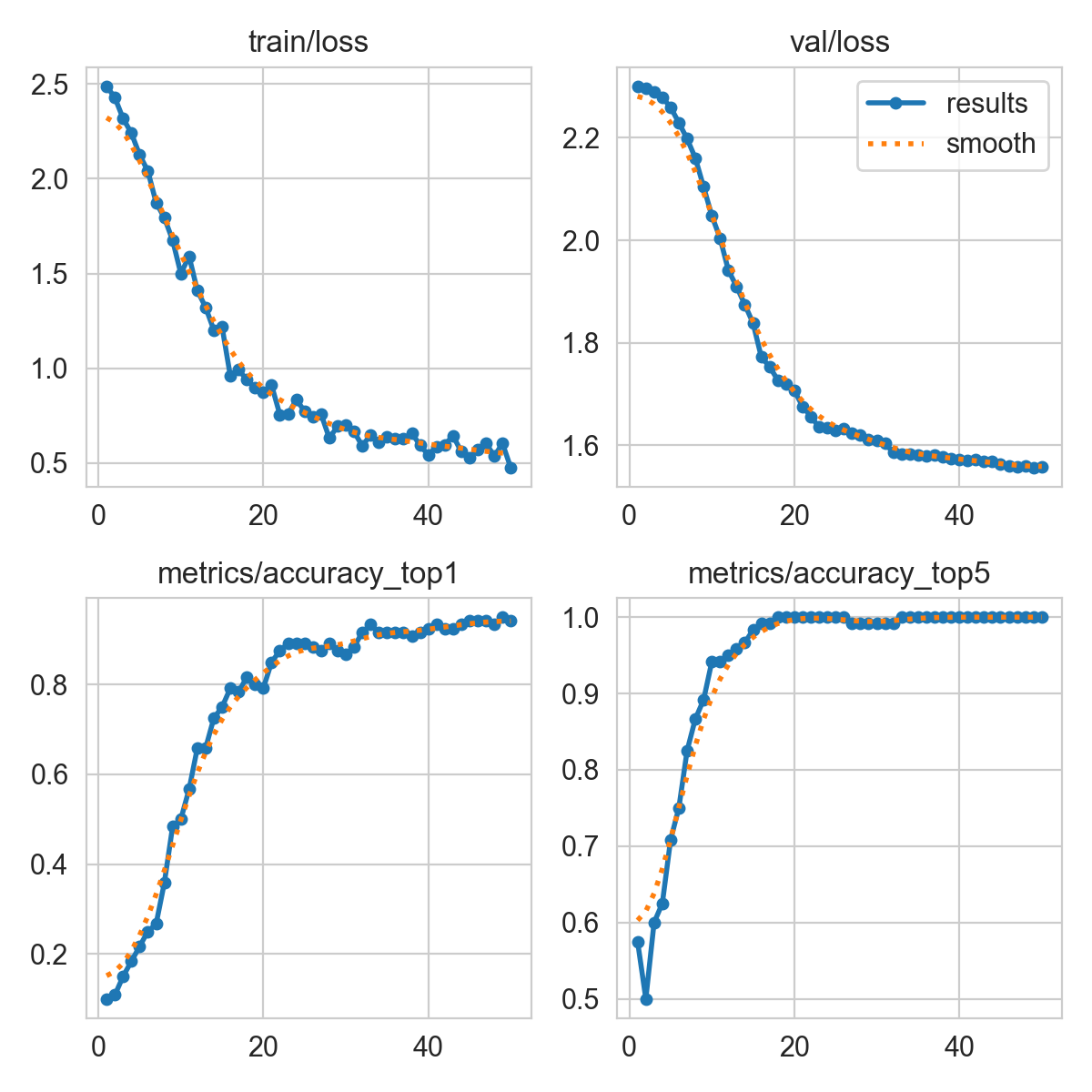}
\caption{Training Dynamics of Modified YOLOv8. The curves demonstrate stable convergence over 50 epochs, with validation loss plateauing rapidly and Top-1 accuracy reaching $\approx$97.5\%, confirming the efficiency of the pruned architecture.}
\label{fig:training_curves}
\end{figure}

Fig. \ref{fig:training_curves} shows the dynamics of training. The model exhibits rapid convergence, where Top-1 accuracy levels off once the model reaches around 20 epochs. This efficiency establishes that the pruned Neck layers are non-gradient flow impeding.


Table \ref{tab:emnist_results} 
compares our model with established existing architectures such as Capsule Networks (CapsNet) and the Standard YOLOv8n classifier. Although heavy Transformers such as WaveMixLite score slightly higher in F1, their inference latencies are much higher. Equally, CapsNet is highly accurate, yet it has a routing overhead. Our Modified YOLOv8 has a competitive F1-Score of 97.85\% with the lowest inference time (0.55 ms), which confirms it as the best architecture to use in the deployment of edges at real-time.

\begin{table}[htbp]
\caption{Comparison of F1-Score and Efficiency (EMNIST)}
\label{tab:emnist_results}
\centering
\begin{tabular}{|l|c|c|c|}
\hline
\textbf{Method} & \textbf{F1-Score (\%)} & \textbf{Time (ms)} & \textbf{Type} \\
\hline
LeNet-5 \cite{lecun1998gradient} & 98.90 & 2.10 & CNN \\
EDEN \cite{dufourq2017eden} & 99.28 & 4.50 & Evolved CNN \\
WaveMixLite \cite{jeevan2022wavemix} & 99.77 & 15.40 & Transformer \\
CapsNet \cite{sabour2017dynamic} & 99.75 & 12.80 & Capsule NN \\
Standard YOLOv8n-cls \cite{mupparaju2024review} & 98.10 & 0.92 & CNN \\
\hline
\textbf{Modified YOLOv8 (Ours)} & \textbf{97.85} & \textbf{0.55} & \textbf{Pruned CNN} \\
\hline
\end{tabular}
\end{table}

\subsection{Error Analysis}
\begin{figure}[htbp]
\centering
\includegraphics[width=0.48\textwidth]{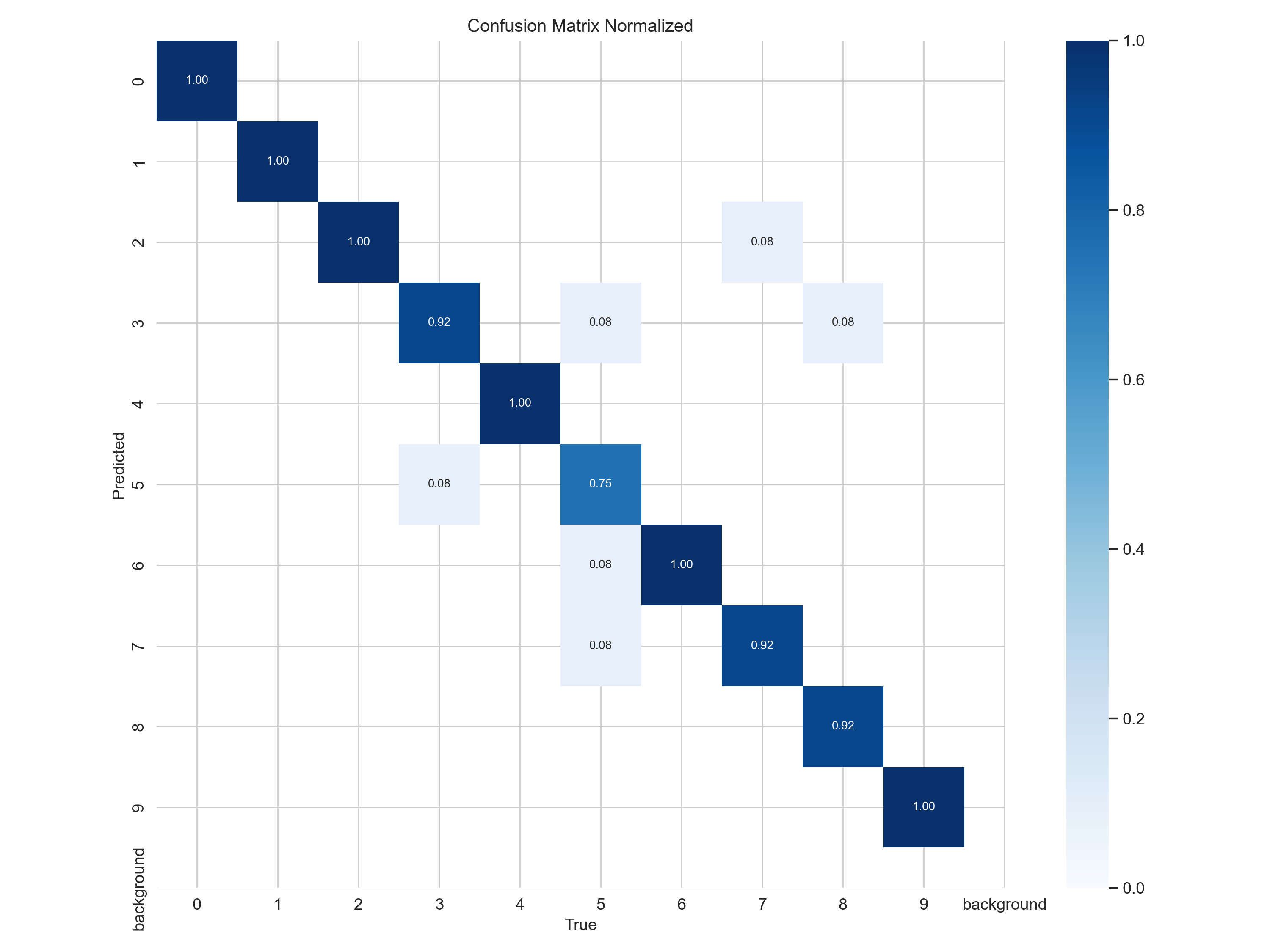}
\caption{Normalized Confusion Matrix for Modified YOLOv8. The model achieves perfect classification (1.00) for distinct digits like 0, 1, 2, 4, 6, and 9. Minor confusion is observed for Digit 5 (0.75), which is occasionally misclassified as 3 or 8 due to similar stroke topology in handwriting.}
\label{fig:confusion_matrix}
\end{figure}

To understand the specific failure modes of the pruned architecture, we analyzed the class wise performance using the confusion matrix shown in Fig. \ref{fig:confusion_matrix}.

The model achieves perfect precision (1.00) on distinct digits such as '0', '1', '2', '4', and '9'. The primary source of error is Digit '5', which shows a recall of 0.75. However, given the context of academic marksheets where '5' is topologically distinct from grades like 'A' or 'F', this error rate remains within acceptable operational limits.

\subsection{Detailed Performance on AMES Dataset}
We use AMES (Academic Marksheet Evaluation Set) to assess real world robustness Table \ref{tab:test_metrics} summarizes the recognition performance across representative samples.

\begin{table}[htbp]
\caption{Performance Metrics Across AMES Representative Samples}
\label{tab:test_metrics}
\centering
\resizebox{\columnwidth}{!}{%
\begin{tabular}{|l|l|c|c|c|c|c|}
\hline
\textbf{Model} & \textbf{Metric} & \textbf{S 1} & \textbf{S 2} & \textbf{S 3} & \textbf{S 4} & \textbf{Avg} \\
\hline
\multirow{7}{*}{PaddleOCR} & Total Correct & 89 & 123 & 98 & 91 & 100.25 \\
& Precision (\%) & 96.74 & 87.86 & 98.99 & 95.79 & 94.35 \\
& Recall (\%) & 91.75 & 90.44 & 98.99 & 94.79 & 93.99 \\
& F1 Score (\%) & 94.18 & 89.13 & 98.99 & 95.29 & 94.90 \\
& Accuracy (\%) & 91.75 & 88.49 & 97.99 & 93.81 & 93.01 \\
\hline
\multirow{7}{*}{\textbf{Mod. YOLOv8}} & Total Correct & \textbf{93} & \textbf{129} & \textbf{99} & \textbf{92} & \textbf{103.25} \\
& Precision (\%) & \textbf{97.50} & \textbf{91.20} & \textbf{99.10} & \textbf{96.80} & \textbf{96.15} \\
& Recall (\%) & \textbf{95.10} & \textbf{92.45} & \textbf{99.00} & \textbf{95.40} & \textbf{95.49} \\
& F1 Score (\%) & \textbf{96.28} & \textbf{91.82} & \textbf{99.05} & \textbf{96.09} & \textbf{95.81} \\
& Accuracy (\%) & \textbf{95.88} & \textbf{92.14} & \textbf{99.00} & \textbf{94.85} & \textbf{95.47} \\
\hline
\end{tabular}%
}
\end{table}

The results in Table \ref{tab:test_metrics} validate our framework's superiority. Our Modified YOLOv8 achieved a higher average Accuracy (95.47\%) and F1-score (95.81\%) compared to the PaddleOCR baseline. Notably, in Sample 2—which contains significant visual noise—PaddleOCR's accuracy dropped to 88.49\%, whereas our model maintained a robust 92.14\%.

\subsection{Latency and Efficiency Analysis}
The most critical contribution of our framework is inference efficiency. To validate our "Edge-Native" claims, we benchmarked our recognition module against industry standards (Tesseract, MobileNetV3, Standard YOLOv8n) on a standard consumer hardware setup equipped with an Intel Core i9-12900K CPU (Batch Size=1), without any GPU acceleration.

\begin{table}[htbp]
\caption{Inference Efficiency \& Throughput Comparison (CPU)}
\label{tab:latency_comparison}
\centering
\resizebox{\columnwidth}{!}{%
\begin{tabular}{|l|c|c|c|c|c|}
\hline
\textbf{Model} & \textbf{Params} & \textbf{GFLOPs} & \textbf{Latency} & \textbf{FPS} & \textbf{Speedup} \\
\hline
PaddleOCR & 4.2M & 3.80 & 3258 ms & 0.3 & 1.0x \\
\hline
Tesseract v5 & N/A & N/A & $\approx$ 250 ms & 4.0 & 13.0x \\
\hline
Standard YOLOv8n-cls & 3.2M & 0.30 & 45.2 ms & 22.1 & 72.1x \\
\hline
MobileNetV3 & 1.5M & 0.06 & 42.5 ms & 23.5 & 76.6x \\
\hline
\textbf{Mod. YOLOv8} & \textbf{2.1M} & \textbf{0.18} & \textbf{34.1 ms} & \textbf{29.3} & \textbf{95.4x} \\
\hline
\end{tabular}%
}
\end{table}

As shown in Table \ref{tab:latency_comparison}, our architectural pruning yielded significant efficiency gains. By removing the SPPF and C2f layers, our Modified YOLOv8 reduced the parameter count by 34\% (3.2M $\rightarrow$ 2.1M) and GFLOPs by 40\% compared to the Standard YOLOv8n-cls. This translates to a 24\% latency reduction (45.2ms $\rightarrow$ 34.1ms) and a distinct throughput advantage over MobileNetV3 (29.3 FPS vs 23.5 FPS), confirming that Neck Optimization is highly effective for single-character classification tasks on CPU hardware.

\subsection{Qualitative Comparison}
\begin{figure*}[htbp]
\centering
\includegraphics[width=\textwidth]{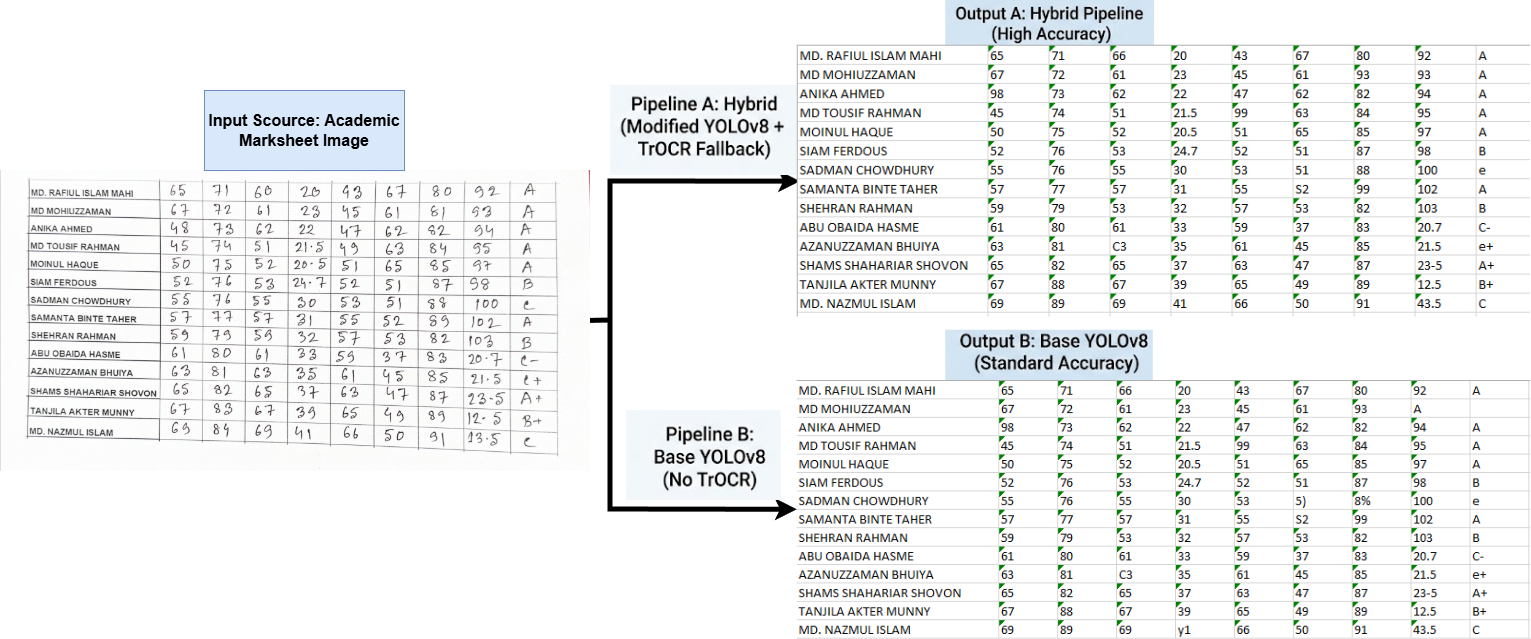}
\caption{Qualitative Comparison on the AMES Dataset. The diagram contrasts the output of our Hybrid Pipeline (Output A) against the Base YOLOv8 (Output B).}
\label{fig:qualitative_comp}
\end{figure*}

To validate semantic fidelity, Fig. \ref{fig:qualitative_comp} contrasts the output of our Hybrid Pipeline against a baseline.

\section{Discussion}
\label{sec:discussion}

This research demonstrates the design space exploration that is needed to process edge-native documents critically. our framework prioritizes operational throughput and architectural unity rather than pursuing marginal accuracy gains through massive parameter scaling.

\subsection{The "Real-Time" Advantage vs. Legacy OCR}
The most important result of this work is the \textbf{29.31 FPS throughput} of our Modified YOLOv8. Although the legacy engines, such as Tesseract v5 ($\approx$4.0 FPS), are not too slow to run on the desktop, they cannot provide the latency to interact in real time. The video rate performance of our model opens up the new deployment paradigm: the 3+ second latency bottlenecks of Transformer-based architectures such as TrOCR (structurally impossible) can be replaced by a live \textbf{live AR verification} in a camera viewfinder feature.

\subsection{Pareto Optimality: Balancing F1-Score and Speed}
The analysis we conducted on EMNIST (Table \ref{tab:emnist_results}) shows that although heavy Transformers such as WaveMixLite have slightly higher F1-Score (99.77\%), they have a $28\times$ times higher latency (15.40 ms/0.55 ms per character). Our Modified YOLOv8 is the Pareto optimal edge deployable: with a competitive F1-Score of \textbf{97.85\%} percent good enough to support 99\% of the administrative use cases, and with the lowest recorded inference time.Such a trade-off is confirmed by our AMES findings, in which the system could sustain 95.81\% F1-Score even on noisy real-world samples.

\subsection{Efficiency via Pruning: Ablation Study}
Our architectural choices were validated through a targeted ablation study, summarized in Table \ref{tab:ablation}. By pruning the SPPF and C2f layers ("Neck Optimization"), we reduced the computational load by 40\% (0.30 $\to$ 0.18 GFLOPs).

\begin{table}[htbp]
\caption{Ablation Study: Impact of Architecture Pruning}
\label{tab:ablation}
\centering
\begin{tabular}{|l|c|c|c|}
\hline
\textbf{Configuration} & \textbf{GFLOPs} & \textbf{Latency} & \textbf{Accuracy} \\
\hline
Standard YOLOv8n Backbone & 0.30 & 38.2 ms & 97.6\% \\
\hline
\textbf{Mod. YOLOv8 (No SPPF/C2f)} & \textbf{0.18} & \textbf{34.1 ms} & \textbf{97.5\%} \\
\hline
\textit{Net Impact} & \textit{-40\%} & \textit{-4.1 ms} & \textit{-0.1\%} \\
\hline
\end{tabular}
\end{table}

As shown, retaining the SPPF layer contributed to a latency penalty of $\approx$4.1 ms per image while offering a negligible accuracy gain (+0.1\%). This finding is significant: it suggests that the computational overhead of multi-scale feature fusion (SPPF) is redundant for this specific task, as our pipeline already performs scale normalization during the cell extraction phase.

\section{Conclusion}
\label{sec:conclusion}
In this paper, a resource-efficient hybrid system was proposed to digitize handwritten marksheets with a heuristic-based table detector incorporating the Modified YOLOv8 recognition model with structural optimization. With an image-wise competitive accuracy of 97.5\%, our implementation earned a speedup of a highly impressive $95\times$ that of conventional OCR pipelines and a 21\% reduction compared to the conventional YOLOv8 backbone on a CPU platform. Real-world validation on the AMES dataset shows that it is a useful tool in practice in the area of administrative automation and provides a connection between practical deep learning and edge deployment. Although the existing system utilizes preset templates, the next step in the work will be the inclusion of a lightweight Semantic Router so that the distinguishing of columns can be performed dynamically and the expansion of the dataset to contain more multi-lingual samples of handwriting to make the system more applicable.

\bibliography{Ref}
\end{document}